\documentclass[letterpaper, 10 pt, conference]{ieeeconf-modified}  %

\IEEEoverridecommandlockouts                              %

\usepackage[numbers]{natbib}
\usepackage{graphicx}
\graphicspath{ {./figures} {../plots/out}}
\usepackage{siunitx}
\usepackage{algorithm}
\usepackage{algorithmic}
\usepackage{amsmath} %
\usepackage{amssymb}  %
\usepackage{textcomp}
\usepackage{hyperref}
\usepackage[capitalize]{cleveref}
\usepackage{makecell}
\usepackage{capt-of}
\usepackage{booktabs}

\crefname{equation}{}{}

\newcommand{\gitlink}{https://aidx-lab.org/manipulation/humanoids24}

\title{\LARGE \bf
Learning Time-Optimal and Speed-Adjustable\\ Tactile In-Hand Manipulation
}
\author{Johannes Pitz \;\;  Lennart Röstel \;\;  Leon Sievers \;\;  Berthold Bäuml%
}

\begin{document}
\twocolumn[{%
        \renewcommand\twocolumn[1][]{#1}%
        \maketitle
        \begin{center}
            \centering
            \vskip -0.8cm
            \includegraphics[width=\textwidth]{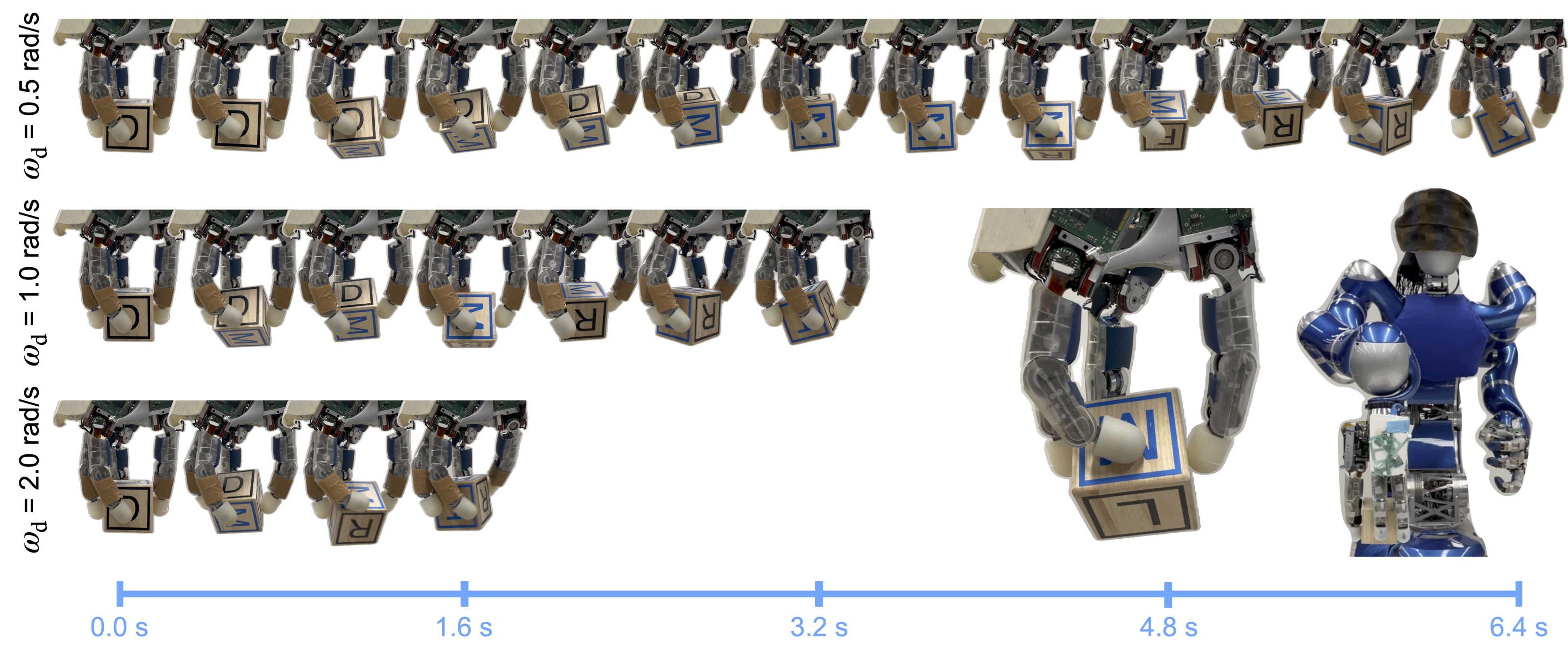}
            \captionof{figure}{The DLR-Hand II \cite{Butterfass2001} is performing the complex task of reorienting a cube to a goal orientation for three different desired orientation speeds spanning a range of factor four (more examples are shown in the accompanying video). 
            The time needed for reorienting the cube matches the desired speed.
            In the lower right, a closeup of the hand and the overall robotic setup with the humanoid Agile Justin is shown. 
            All reorientations are performed purely tactile, using only the hand's position and torque sensors (no visual input, hence the blindfolded robot).}
            \label{fig:title_figure}
        \end{center}%
    }]

{\let\thefootnote\relax\footnote[0]{
    \newline The authors are with the Learning AI for Dextrous Robots Lab, Technical University of Munich, Germany (\href{https://aidx-lab.org/}{\scriptsize\texttt{aidx-lab.org}}), and the DLR Institute of Robotics \& Mechatronics (German Aerospace Center). \newline
Contact:{\tt\scriptsize{ \{johannes.pitz|berthold.baeuml\}@tum.de}}}}

\thispagestyle{empty}
\pagestyle{empty}

\begin{abstract}
    In-hand manipulation with multi-fingered hands is a challenging problem that recently became feasible with the advent of deep reinforcement learning methods. While most contributions to the task brought improvements in robustness and generalization, this paper addresses the critical performance measure of the speed at which an in-hand manipulation can be performed. We present reinforcement learning policies that can perform in-hand reorientation significantly faster than previous approaches for the complex setting of goal-conditioned reorientation in SO(3) with permanent force closure and tactile feedback only (i.e., using the hand's torque and position sensors). Moreover, we show how policies can be trained to be speed-adjustable, allowing for setting the average orientation speed of the manipulated object during deployment. To this end, we present suitable and minimalistic reinforcement learning objectives for time-optimal and speed-adjustable in-hand manipulation, as well as an analysis based on extensive experiments in simulation. We also demonstrate the zero-shot transfer of the learned policies to the real DLR-Hand II with a wide range of target speeds and the fastest dextrous in-hand manipulation without visual inputs.
     \\
    Website: \href{\gitlink}{\scriptsize\texttt{\gitlink}}
\end{abstract}

\section{Introduction}
In-hand manipulation has the potential to provide an efficient way for reorienting objects from one configuration to another. 
While traditional approaches using manipulators with grippers often require intermediate placement on external surfaces for re-grasping, in-hand reorientation allows for more direct solutions. 
The advantage of in-hand manipulation thus hinges on the ability to perform the reorientation quickly and reliably, where faster execution times can translate directly to increased productivity in industrial applications. 

Motivated by this, we present progress toward fast and reliable in-hand reorientation. 
Here, we only experiment with simple cuboids because the faster training times facilitate a more detailed analysis.
However, we expect the results to hold for in-hand manipulation of any objects and apply to other robotic tasks that can be solved with reinforcement learning.

We consider two objectives: First, we study the case of time-optimal in-hand manipulation, where we demonstrate the capability to perform goal-conditioned in-hand reorientations several times faster than any prior approach while retaining high success rates ($>95\%$ in simulation).
Then, we propose methods for learning in-hand reorientation policies with adjustable reorientation speeds, allowing us to set variable speeds at deployment without retraining.

Rotating objects at desired speeds could reduce the wear and tear of robotic hands by making slower movements, for example, in an assembly line that runs at reduced speed for various reasons.
However, we are mainly interested in finding out if it is possible to solve this complex problem with reinforcement learning and to analyze the necessary modifications of the baseline method.
Moreover, these modifications could also be analyzed in the context of adjusting the speed in legged locomotion or other tasks that need to be conditioned similarly.

\subsection{Related Work}
In a line of work on in-hand manipulation with high-speed visual sensing~\cite{Shiokata2005-lg, Furukawa2006-dc,Higo2018-qc}, Furukawa et al. show the impressive skill of dynamically re-grasping a cylindrically-shaped object by throwing and catching. 
While the system allows extremely fast reorientations, the approach seems difficult to generalize to different objects and target orientations, to execute reliably~(\citet{Furukawa2006-dc} report 35\% success rate), and requires specialized hardware.

\citet{Morgan2022-wr} propose a planning-based approach for in-hand reorientation with finger gaiting, where the object pose is tracked via a visual pose estimator. 
However, partly due to the need for online re-planning, the method has limitations in terms of execution speed, requiring up to \SI{153}{s} to perform a single reorientation.

Recently, learning-based methods have shown promising advancements in the domain of in-hand manipulation~\cite{OpenAI2018-yi,openai2019rubiks}. 
While classical methods are traditionally challenged by complex multi-contact dynamics and high dimensional state and control spaces, reinforcement learning (RL) has shown great potential for deriving efficient and robust controllers under these conditions.

In particular, the sim2real approach, where RL policies are trained in simulation and then transferred to the real world, has gained considerable attention~\cite{Sievers2022,Pitz2023-ra,Rostel2023-nc,Chen2023-mq,Qi2023-th}.

Several prior works train policies for reorienting objects using supporting surfaces (including the palm of the hand)~\cite{OpenAI2018-yi, Handa2022-rc,Yin2023-tw,Yuan23}.
However, this raises the need for additional (wrist) movements when deployed as part of a pick-and-place pipeline, leading to an increase in end-to-end execution time.

Consequently, there is a growing interest in the (more challenging) setting where the hand is pointed downwards during in-hand manipulation, requiring permanent force closure between the fingers and the object \cite{Sievers2022,Pitz2023-ra,Rostel2023-nc,Chen2023-mq}.
In this setting,~\citet{Chen2023-mq} show the reorientation of diverse, complex objects with a single policy using depth sensing of an external camera, reporting a median reorientation time of~$\sim$~\SI{7}{s}.

Finally, many of the previously mentioned works consider the setting where visual feedback from cameras is available for tracking the object pose~\cite{Furukawa2006-dc,OpenAI2018-yi,Chen2023-mq,Handa2022-rc,Qi2023-th,Morgan2022-wr,Yin2023-tw}. This assumption may constrain the applicability of the methods in practice.
Hence, prior work on purely tactile in-hand manipulation~\cite{Sievers2022, Pitz2023-ra, Rostel2023-nc,Pitz2024-shape}, learn policies based on proprioceptive (joints' angles and torque) feedback alone. 
In this regime,~\citet{Pitz2024-shape} show autonomous in-hand reorientation with generalization to novel geometric objects with a mean reorientation time of~$\sim$~\SI{4}{s}.

\subsection{Contributions}
\label{sec:contributions}
The research presented in this paper builds upon prior work on in-hand manipulation without visual sensing, using position and torque measurements as the only sensor modalities~\cite{Sievers2022, Pitz2023-ra, Rostel2023-nc,Pitz2024-shape}. 
We train policies to perform reorientation with the hand facing downwards and also consider the practically relevant task of goal-conditioned in-hand reorientation, where the policy autonomously reorients the object to given target orientations in $\mathrm{SO}(3)$. 

The main contributions presented in this work are:
\begin{itemize}
    \item Methods for learning in-hand manipulation policies that are \textit{speed-adjustable} with a detailed analysis of suitable reinforcement learning objectives.
    \item Pushing boundaries regarding \textit{fast} in-hand reorientation while maintaining high success rates.
    \item Experiments on the real DLR-Hand~II, performing the fastest ever shown robotic system for $\mathrm{SO}(3)$-general in-hand reorientations, close to human capabilities in terms of speed.
\end{itemize}

\section{System Overview}
\label{sec:system_overview}
In the following, we provide an overview of the system used for in-hand reorientation in this work.
We first briefly describe the hardware \ref{sec:hardware} and control architecture \ref{sec:control_architecture} used. 
In line with prior work~\cite{Pitz2023-ra,Rostel2023-nc,Pitz2024-shape}, we then introduce two learned components for achieving the in-hand manipulation task itself: 
an RL control policy\ref{sec:policy_controller} and a learned, purely tactile state estimator~\ref{sec:state_estimator}.
\subsection{Hardware}
\label{sec:hardware}
We use the DLR-Hand~II, a 4-fingered robotic hand with 16 joints and 12 actuated degrees of freedom (DOF) with joint torque sensing~\cite{Butterfass2001}.
The hand is highly performant regarding speed and torque (max. joint velocity $\dot{q}_{\text{max}}\approx \SI{9.6}{rad/s}$, max. joint acceleration $\ddot{q}_{\text{max}}\approx \SI{110}{rad/s^2}$) and measurement accuracy, allowing for high-fidelity impedance control.
We use calibrated kinematic parameters of the hand as obtained by the calibration procedure described by \citet{Tenhumberg2023-Hand}.

\subsection{Control Architecture}
\label{sec:control_architecture}
We employ a hierarchical control architecture, where a low-level impedance controller is used to track desired joint targets from an RL policy~\cite{Sievers2022}.
This allows running the neural network policies and state estimators at lower frequencies $f_{\text{nn}}$.
In our setup, policy actions $a$ are first-order low-pass filtered with a time constant $\tau$ and given as desired joint targets $q_\text{d}$ to the impedance controller at a rate of $f_{\text{pd}}=\SI{1}{kHz}$. 
Contacts can be implicitly detected by monitoring the difference between measured joint positions $q$ and desired joint targets $q_\text{d}$~\cite{Rostel2022-gu}.
Measured joint angles $q\in \mathbb{R}^{12}$ and desired joint angles $q_\text{d}\in \mathbb{R}^{12}$ are the only inputs to our learned system during reorientation. 

During preliminary experiments, we found out that to achieve a significant speed-up for the in-hand manipulation, we had to increase the interaction frequency $f_{\text{nn}}$ from typically $\SI{10}{Hz}$ to $\SI{20}{Hz}$ and reduce $\tau$ from $\SI{0.5}{s}$ to $\SI{0.2}{s}$.
Even higher $f_{\text{nn}}$ in combination with lower $\tau$ showed slightly faster results in simulation, but we chose $\SI{20}{Hz}$ for practical reasons like the network latency. 
Note that the magnitude gain for the respective network interaction frequency stays comparable.

\begin{figure}
    \centering
    \vspace{2mm}
    \includegraphics[width=0.95\linewidth]{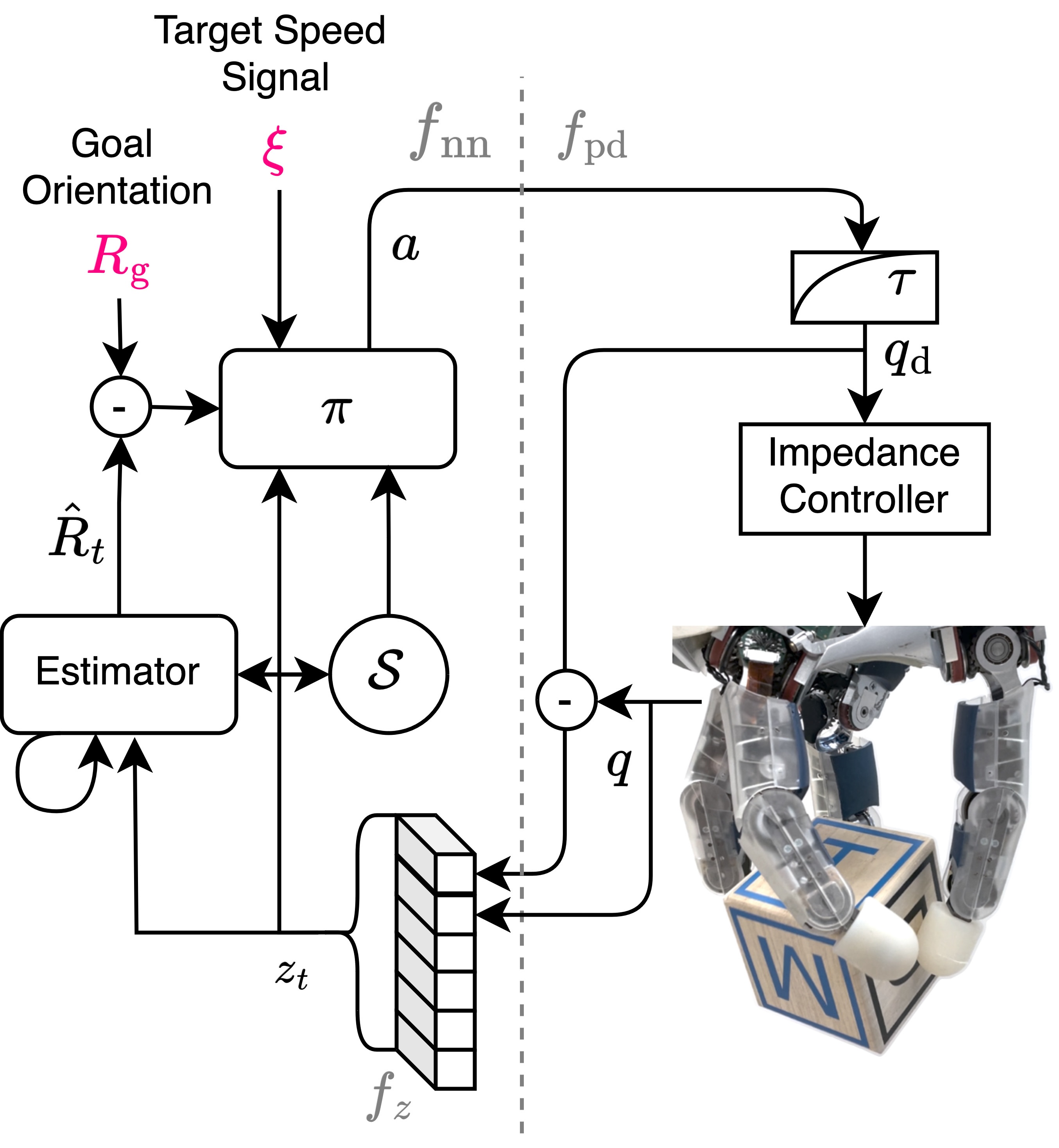}
    \caption{Overview of control architecture and system components. 
    We use a learned state estimator $\rho$ to estimate the object pose $\hat{s}_t$ from proprioceptive (joints' torques and angles) observations $z_t$. 
    Based on the estimated state, a shape encoding $\mathcal{S}$ is computed and used as input to the control policy $\pi$.
    The control policy is additionally conditioned on the desired object orientation $R_\text{g}$ and optionally a target speed signal $\xi$, which controls the speed of reorientation.
    The actions of the policy are low-pass filtered and given to an underlying impedance controller for the torque-controlled DLR-Hand~II.
    }
    \label{fig:overview}    
\end{figure}

\subsection{State Estimation}
\label{sec:state_estimator}

We use a learned state estimator $\rho$ for estimating the pose of the object 
$$\hat{s}_t = (\hat{x}, \hat{R})_t \in \mathrm{SE}(3)$$
with position $\hat{x}$ and orientation $\hat{R}$ at each time step $t$. 
The state estimator is trained in a realistic simulation to recurrently predict the object pose from the history of proprioceptive observations $z_{0:t}$ as proposed in~\citet{Rostel2023-nc}.
Here, each observation $z_t$ is a stack of $k=6$ tuples
\begin{align}
    z_t = \left[ q(t-t_i), e(t-t_i)\right]_{i = 0:(k-1)}, \quad t_i = \frac{i}{f_z},
\end{align}
with measured joint angles $q$ and control error $e = q_\text{d} - q$, sampled at a frequency of $f_{z}=\SI{60}{Hz}$, corresponding to a time window of $\SI{0.1}{s}$.
Additionally, the state estimator is conditioned on the object mesh, from which we compute a shape encoding $\mathcal{S}(\hat{x}_t, \hat{R}_t)$.
Precisely, this encoding is the set of vectors from a number of fixed ``basis points''~\cite{Prokudin2019-sg} to the surface of the mesh transformed to the estimated pose.
The estimated state is also input to the control policy during training and deployment. 
We refer to~\citet{Pitz2024-shape} for a detailed description of the architecture. 

\subsection{Policy Controller}
\label{sec:policy_controller}
We train a goal-conditioned policy
\begin{align}
    a_t = \pi_{\varphi}\left(z_t, R_\text{g}\hat{R}^{-1}_t, \mathcal{S}(\hat{x}_t, \hat{R}_t), \xi\right)
\end{align}
to perform the task of in-hand reorientation (see \cref{sec:rl_task}).
The input to the policy is the concatenation of the stack of proprioceptive observations $z_t$, the estimated relative rotation to the goal $R_\text{g}\hat{R}^{-1}_t$ and the shape encoding based on the estimated object pose. 

In \cref{sec:variable_speed}, we additionally propose to condition the policy on a ``target speed'' signal $\xi$ as an additional input, which enables controlling the reorientation rate during deployment.

We optimize the learnable parameters $\varphi$ of the policy, parametrized as a feedforward neural network, by Proximal Policy Optimization (PPO)~\cite{Schulman2017proximal}. 
To obtain control strategies that transfer robustly to the real system, we employ the Estimator-Coupled Reinforcement Learning (EcRL) training scheme~\cite{Rostel2023-nc}, 
where the policy is trained to become robust to the state estimator's inaccuracies and ambiguities in simulations with noise and domain randomization. 
For training in simulation, we use the GPU-accelerated rigid-body simulator Isaac Sim \cite{IsaacSim} with a realistic parameterization of contacts and shared experience buffers for policy and estimator.

\begin{figure*}
    \centering
    \vspace{2mm}
    \includegraphics{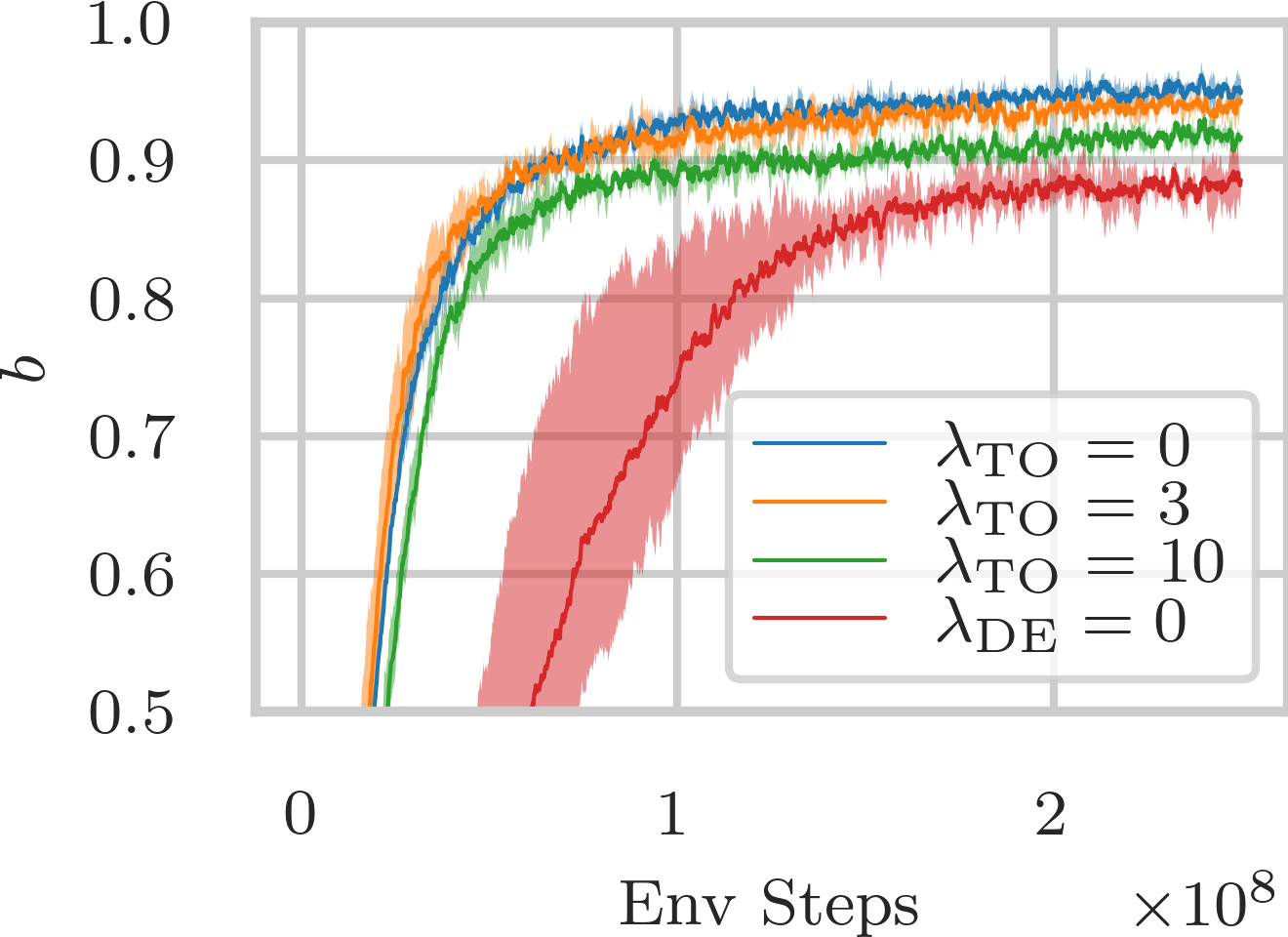}
    \includegraphics{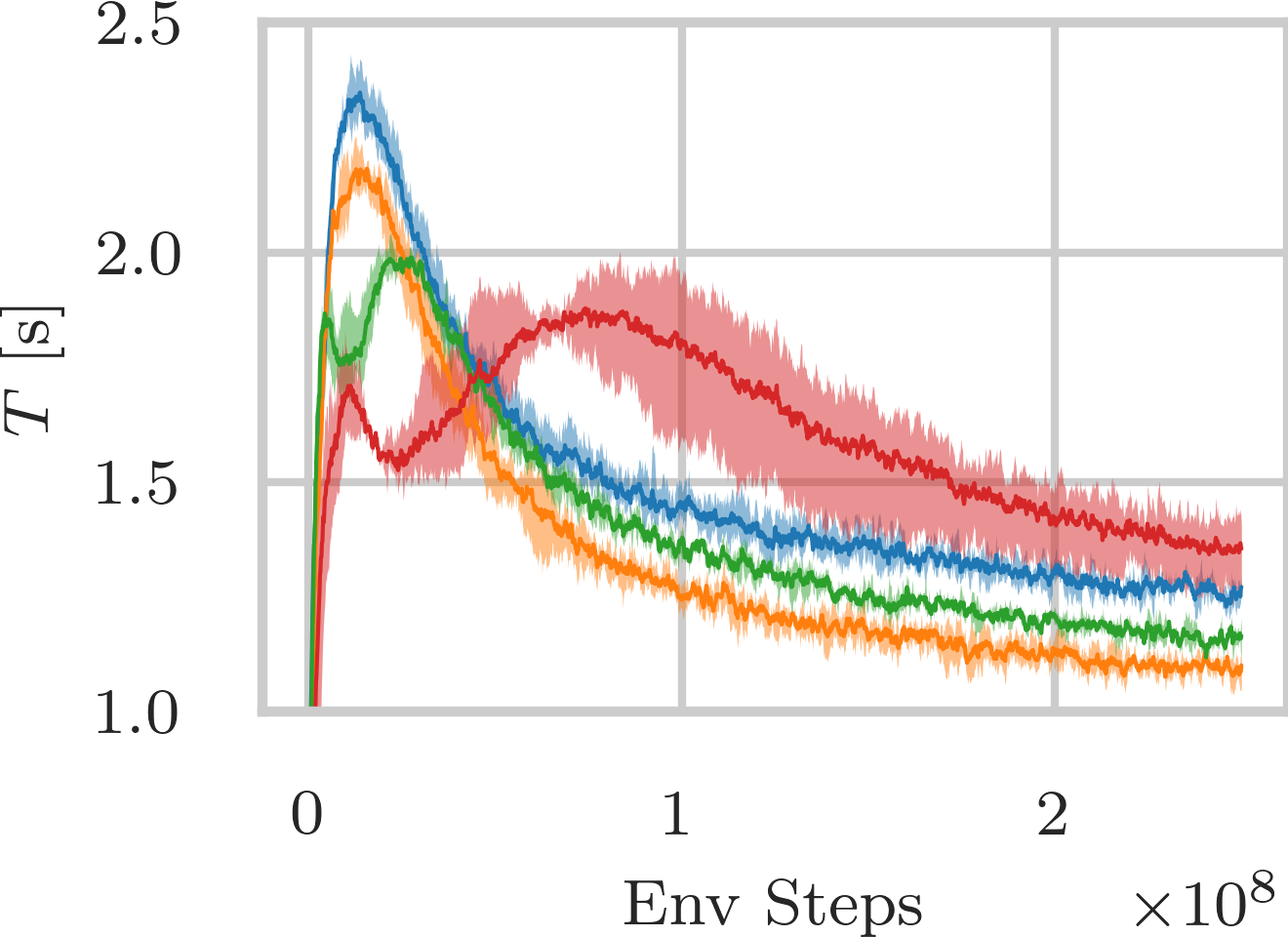}
    \includegraphics{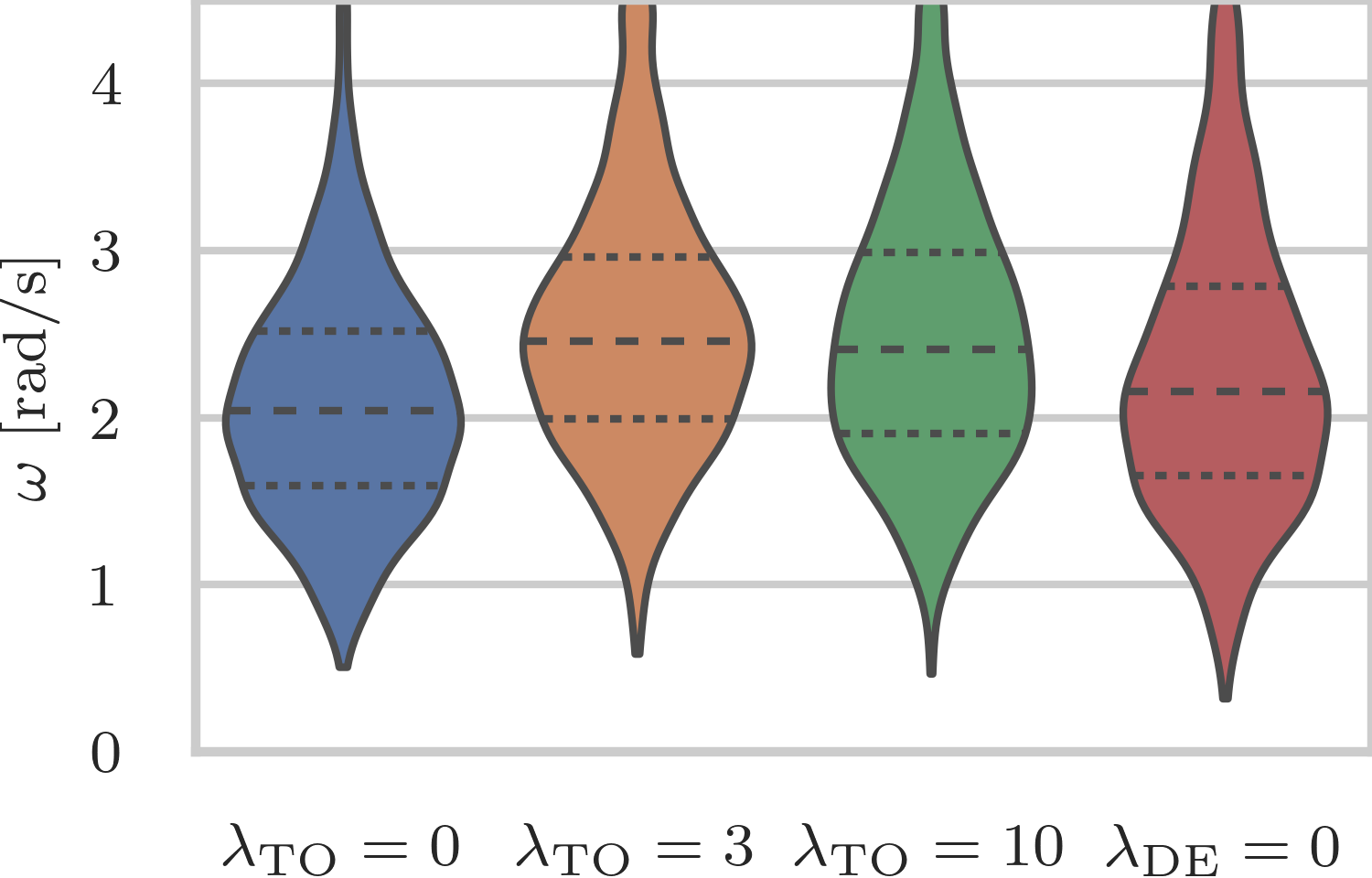}
    \caption{
        (Left) Success rate $b$ and (center) average time $\mathrm{T}$ required to reach the first goal are plotted over the training progress. Each line is the mean over three training runs, with shaded areas covering the min and max. We smooth the signal of the individual runs.
        (Right) Average angular velocity $\omega = \theta_0/T$ over evaluation episodes. We ran 1200 episodes with a single policy each and discarded failed episodes ($<$ 3\%) and episodes where $\theta_0 < \pi / 4$ to avoid high variance due to small numbers and reorientations without regrasping.  
    }
    \label{fig:exp_bonus}    
\end{figure*}

\subsection{Reinforcement Learning Task}
\label{sec:rl_task}
In-hand reorientation is the task of bringing an object from an initial orientation $R_0\in \mathrm{SO}(3)$ to a desired goal orientation $R_\text{g}\in \mathrm{SO}(3)$ by in-hand manipulation.
We assume the object to be in a stable grasping state at $t=0$. 
The task is considered successful if the angle $\theta_H = d(R_\text{g}, R_H)$ between the target orientation and the actual object orientation at the final timestep $H$ is smaller than a threshold $\Delta_g = 0.4$rad.

The horizon $H$ is either fixed at 5\,s, or varied according to the initial angle to the goal orientation $\theta_0$ as discussed in \cref{sec:variable_speed}.

During training, if the object is (re-)orientated according to the goal at step $H$, a new goal $R_\text{g}$ is sampled, and the episode continues. 
This trains the policy to perform multiple consecutive reorientations and in practice, leads to well-controlled grasps at the end of each reorientation.
If the object is not orientated correctly or drops during the episode, the environment is reset.
A timeout truncates episodes to ensure new domain randomization samples~\cite{Rostel2023-nc}.
The objects used for training and evaluation in simulation are cuboids with aspect ratios of up to two.
During evaluation, we do not sample consecutive goals but instead reset the environment, and we ignore the target time~$T_\text{d}$ requirement for determining success.
Instead, we report the time to reach the goal~$T$ or the effective speed $\omega = \theta_0 / T$ of as the first timestep when the angle $\theta_t$ was sufficiently small.

The policy $\pi$ is trained to maximize objective $J(\pi)$, which is the expected sum of returns with discount factor $\gamma$ over trajectories $\tau$ in simulation:
\begin{equation}
 J(\pi) = \mathbb{E}_{\tau \sim \pi} \left[ \sum_{t} \gamma^t r_t \right].
\end{equation}
The exact reward function used varies in the following sections. 
Therefore, we describe them there.

\section{Time-Optimal In-Hand Reorientation}
\label{sec:time_optimal}

The time-optimal objective for in-hand reorientation entails minimizing the time required to reach the goal orientation.
To encourage genuinely time-optimal behavior, we can reward the policy if and only if the object is currently in the goal orientation:
\begin{align}
    \label{eq:reward_to}
    r_t^{\text{TO}} =  \begin{cases}
        \lambda_\text{s} \quad &\text{if} \quad d(R_\text{g}, R_t)<\Delta_g \\
        0 \quad &\text{else}.
        \end{cases}
\end{align}
However, we found that this reward signal alone leads to policies that are inferior in terms of success rate and reorientation speed compared to policies trained on a dense reward similar to our previous work (i.e., \cite{Pitz2024-shape}).
\begin{align}
    \label{eq:reward_de}
    r_t^{\text{DE}} = \quad\; &\lambda_{\theta} \left(\theta_{t-1} - \theta_{t} \right) + r_t^\text{HE}, \quad \text{with} \\
r_t^\text{HE} = \quad\; &\lambda_{x} \left( \Delta{x}_{t-1}- \Delta{x}_t \right) \\
 - &\lambda_{q}\; \| ( q_t - \bar{q}_0 )^4 \|_1 \nonumber
\end{align} and coefficients  $\lambda_{\theta}= 1$, $\lambda_{x}= 8$, $\lambda_{q} = \frac{1}{24}$.
Here, $\theta_{t-1} - \theta_{t}$ is the difference in angles to the goal orientations between the previous and current timesteps,
$\Delta{x}_t = ||x_t - x_0||_2$ is the distance of the object position to the initial position, and $\bar{q}_0$ is the mean initial joint configuration.

In contrast to \citet{Pitz2024-shape}, we do not clip the angle reward and use less weight on the joint penalty.
Previously, we have used clipping (cf.~\cref{eq:reward_cl}) to remove the incentive for the policy to make fast and erratic movements. 
We assumed they wouldn't allow proper sim2real transfer and, in particular, make the estimation problem more difficult. 
However, this assumption is likely unnecessary since we started training the estimator along the policy (EcRL)~\cite{Rostel2023-nc}. 
Therefore, we first investigate how much incentive the policy needs to optimize for speed in the oracle setting (without training an estimator). Later, we see how training with the estimator slows down the policy~(cf.~\cref{fig:scatter_speed}).

The dense reward function \cref{eq:reward_de} does not directly incentivize fast reorientation because the available positive reward is limited by the current deviation from the goal orientation (and position).
However, two factors encourage speed.
Firstly, the discount constant $\gamma$ prioritizes immediate rewards over future ones.
And secondly, since the policy does not observe time, it ``fears" at every step that the episode will be terminated soon.
That means the reward available for the remaining deviation and the reward from further new goals could be lost.

Therefore, we analyze how close the dense reward function is to the time-optimal objective by training linear combinations of the two reward functions:
\begin{equation}
    r_t = \lambda_{\text{DE}} \; r_t^\text{DE} + \lambda_{\text{TO}} \; r_t^\text{TO}.
\end{equation}

We compare the sparse reward configuration $\lambda_{\text{DE}}=0$ (where $\lambda_{\text{TO}}=1$) and three variants with dense reward ($\lambda_{\text{DE}}=1$). 
With $\lambda_{\text{TO}}=0$, we have the reward function that we used in \citet{Pitz2024-shape} but without the clipping term~\cref{eq:reward_de}.
And we set $\lambda_\text{s} = 0.03$ for all experiments, such that with the configurations $\lambda_{\text{TO}}=3$ and $\lambda_{\text{TO}}=10$ the bonus contributes around three or respectively ten times more than the relative angle reward (which in turn dominates the other dense components).

In \cref{fig:exp_bonus}, we show the success rate $b$ and the average time $\mathrm{T}$ required to reach the first goal during training and the distribution of the average angular velocity $\omega = \theta_0/T$ of evaluation episodes.
All configurations learn successfully, but $\lambda_{\text{TO}}=10$ and $\lambda_{\text{DE}}=0$ show some irregularities compared to the other two.
The center plot shows a dip in time $\mathrm{T}$, likely caused by the policy prioritizing quick successes over general skills.
Also, the more dominant the sparse reward is, other metrics, such as the average final distance to target $x_H$ or the joint penalty reward, are higher (i.e., worse).
However, these differences do not yield faster policies, as seen in the center and the right plot.
The fastest policies come from the $\lambda_{\text{TO}}=3$ configuration which average around $\SI{2.5}{rad/s}$ while the $\lambda_{\text{TO}}=0$ configuration averages around $\SI{2.0}{rad/s}$.

Since larger $\lambda_{\text{TO}}$, which should yield more time-optimal policies, does not improve the results, we conclude that the reorientation speed is limited at around $\SI{2.5}{rad/s}$ in the oracle setting due to the simulated hardware constraints, the interaction frequency, and filter constant that we are using.
The fact that we can get close to that speed with our base reward is interesting, but the gap shows that the incentives to optimize for speed can still be increased.

\section{Fixed-Speed In-Hand Reorientation}
\label{sec:fixed_speed}

\begin{table}
    \centering
    \vspace{3mm}
    \caption{Fixed speed with various horizons}
    \begin{tabular}{l c c c c c c }
        \toprule
        \multicolumn{1}{c}{$\omega_\text{d}$} [rad/s] & \multicolumn{3}{c}{1.5} &\multicolumn{3}{c}{0.75} \\
        $H_\text{exp}$ [s]& 0.5 & 2.0 & 5.0 & 0.5 & 2.0 & 5.0 \\
        \midrule
        $\mu(\omega)$ [rad/s] & 1.64 & 1.37 & 1.18 & 1.02 & 0.85 & 0.75 \\
        $\sigma(\omega)$ [rad/s] & 0.29 & 0.29 & 0.26 & 0.18 & 0.17 & 0.16 \\
        \bottomrule
    \end{tabular}
    \label{tab:exp_time}
\end{table}

\begin{figure*}
    \centering
    \vspace{2mm}
    \includegraphics{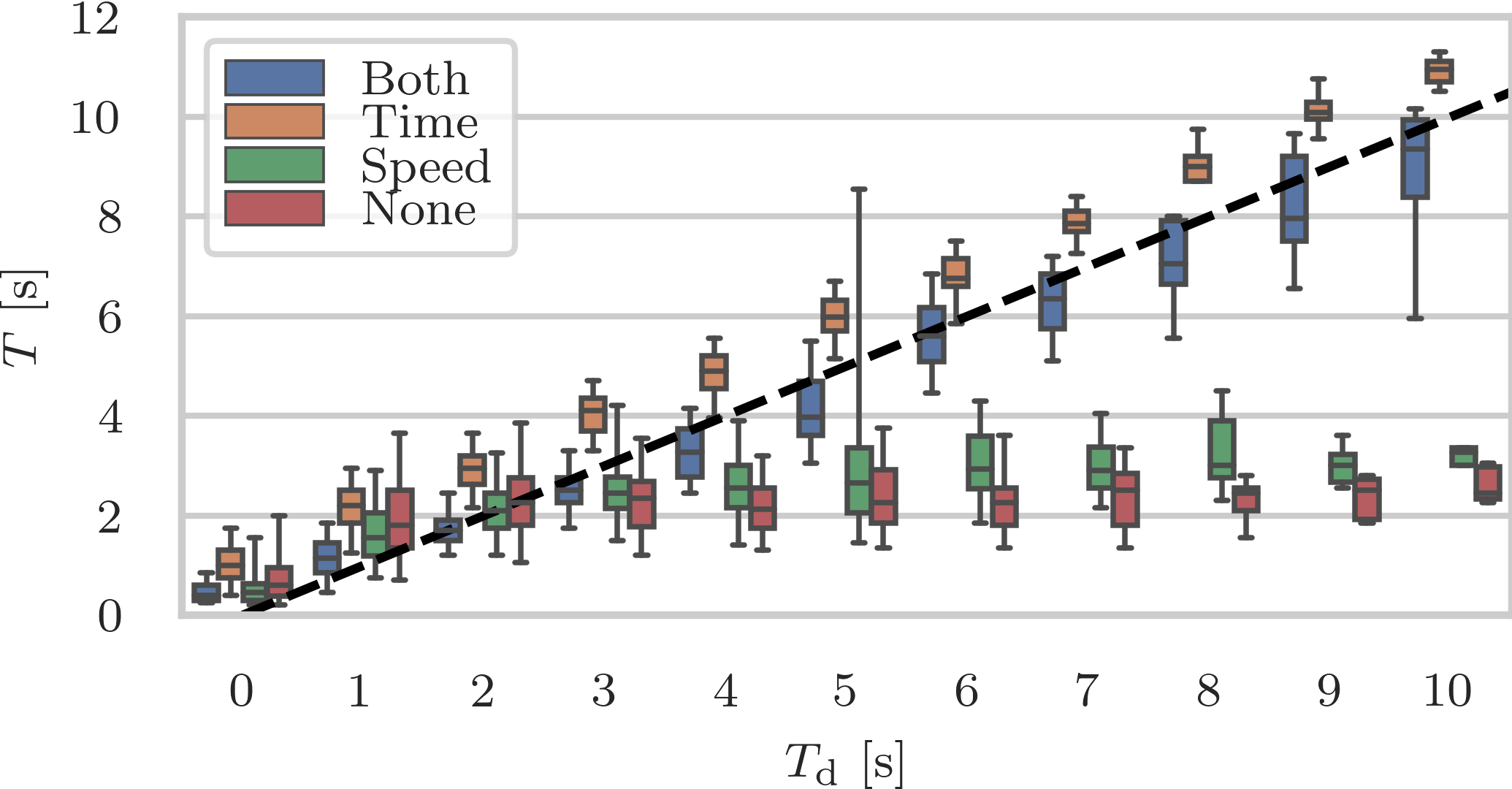}
    \includegraphics{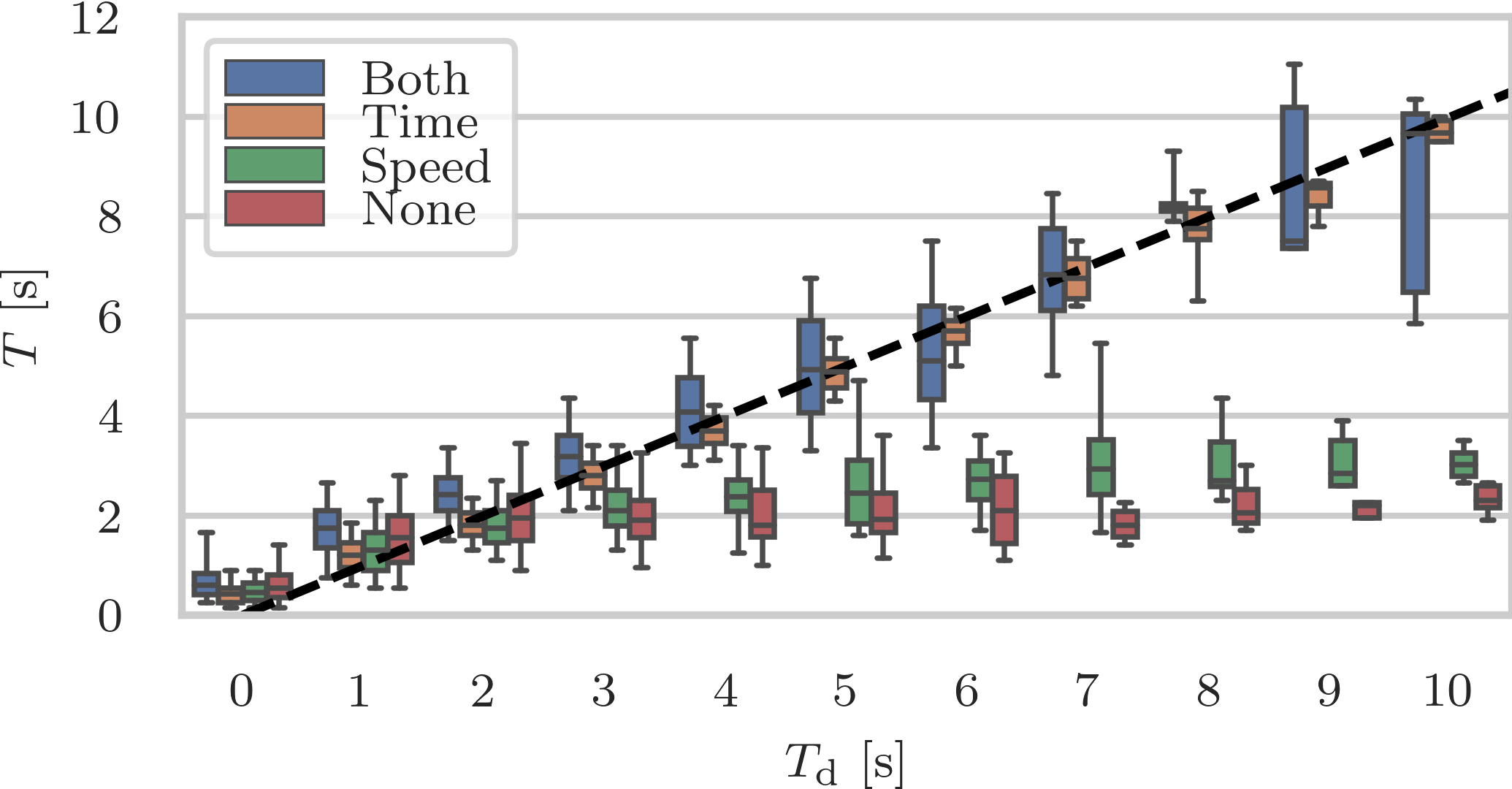}
    \caption{
        Box plot of the time to reach the goal $T$ of evaluation episodes grouped by the target time $T_\text{d}$ rounded to the nearest integer.
        We ran 1200 episodes with a single policy each and discarded failed episodes ($<$ 3\%).
        Whiskers indicate the 5th and 95th percentile.
        (Left) $H_\text{exp} = 2$\,s. 
        (Right) $H_\text{exp}$~is sampled between 0 and 1\,s.
    }
    \label{fig:box_speed}    
\end{figure*}

To achieve fixed-speed reorientations, we considered reward functions based on velocity matching similar to that used with quadruped robots \cite{miki2022learning} but with no success.
Therefore, we reintroduce the clipping operation, which we used in previous work to slow the policy down and count on the incentives explained in \cref{sec:time_optimal} to speed the policy up to match the desired speeds.
We cannot use the time-optimal reward $r^{\text{TO}}$ because that would overshadow the clipping.
So, the full reward function of this section is:
\begin{align}
    \label{eq:reward_cl}
    r_t^{\text{CL}} = \lambda_{\theta} \min \left( \theta_{t-1} - \theta_{t}, \theta_\text{clip} \right) + r_t^\text{HE}.
\end{align}
Of the other factors that incentivize speed, we are mainly interested in the horizon to control the speed of the rotations since the discount factor~$\gamma$ is essential to encourage the policy to optimize for long-term success, which for in-hand manipulation means stable grasps at the goal orientation to prepare for new goals.

To better understand the effects of the horizon, we train policies for two different target speeds $\omega_\text{d}$ and resulting clipping values $\theta_\text{clip}$ and three variants for the horizon $H$.
We set the horizon individually for each episode based on the initial angle to the goal and a constant depending on the experiment $H = \theta_0 / \omega_\text{d} + H_\text{exp}$.

We report the mean and standard deviation over the average angular velocity $\omega = \theta_0/T$ over evaluation episodes in \cref{tab:exp_time}. Just as in \cref{fig:exp_bonus}, we ran 1200 episodes with a single policy each and discarded failed episodes ($<$ 5\%) and episodes where $\theta_0 < \pi / 2$ to avoid high variance due to small numbers and reorientations without regrasping.  

There are a few interesting effects to see in the table.
Firstly, short horizons push the policy to exceed the target speed.
Here, the ``fear" of missing the next episode seems to outweigh the lost reward due to the clipping.
Secondly, the more time the policy has, the slower it gets.
For $\omega_\text{d} = 1.5$ and $H_\text{exp} = 5$, it slows down significantly below the target speed, probably trying not to run into the clipping and losing additional time during finger-gaiting phases where the object is not moving.
But maybe most surprisingly, for the slower target speed $\omega_\text{d} = 0.75$, even $H_\text{exp} = 2$ is insufficient for the policy to slow down below the clipping range.
Again, this must be because missing a bit of reward along the way is outweighed by the chance that the episode end is closer than expected (based on the current angle $\theta_t$).
The experiments in the next section support this explanation by showing that if the policy receives the horizon as an observation, it prefers to use the additional time.

\section{Learning Speed-Adjustable In-Hand Reorientation}
\label{sec:variable_speed}

To allow speed-adjustable reorientation, we use the reward function with angle clipping~\cref{eq:reward_cl} but now pass the policy information about the target speed $\omega_\text{d}$. 
In \cref{fig:overview}, we indicate this additional observation as $\xi$.
We evaluate two possible signals.
\begin{enumerate}
    \item The target speed $\xi=\omega_\text{d}$ directly, which is also used for the angle clipping operation.
    \item The remaining time $\xi= T_\text{d} - t$ to the target time $T_\text{d} = \theta_0 / \omega_\text{d}$ (and note: $H = T_\text{d} + H_\text{exp}$).
\end{enumerate}
Neither signal alone makes the problem fully observable. 
However, we have always worked with partial observability (the policy does not see the horizon).
Therefore, we run experiments with both signals, either one or none, for reference.

In \cref{fig:box_speed}, we show two experiments. 
On the left side, $H_\text{exp} = 2$\,s is fixed, while on the right, we sample $H_\text{exp}$ between 0 and  1\,s (making the problem partially observable again even with both observations).
For both experiments we sample the target speed $\omega_\text{d}$ from $0.25$ to $\SI{2.5}{rad/s}$.

As expected, the policy trained with no target speed signal cannot adapt. 
It learned in both experiments to reorient the objects as fast as possible.
Interestingly, the policies which were passed the target speed directly also didn't adapt.
The policies that see the remaining time show a strong correlation to the requested target time.
Surprisingly, the ones that see only the horizon have a smaller variance than those that see both.
We do not have a good explanation for why this is the case.
Still, we assume that with different network configurations and hyperparameters, the ``Both" policies should perform at least as well as the ``Time" policies.

On the left, we can see that the ``Time" policy consistently takes more time than the original target, indicating that it learned to exploit all the available time ($H_\text{exp} = 2$\,s) even for slow target speeds, which is contrary to the slower fixed-speed policy in \cref{sec:fixed_speed}.
On the right, the ``Time" policy does what we tried to accomplish by being just slightly faster than the respective target time. 

Therefore, we use that setting, the ``Time" policy with varying $H_\text{exp}$ between 0 and 1\,s, to train an RL agent with estimator as described in \cref{sec:system_overview}.

\subsection{Evaluation in Simulation}
\label{sec:estimator_sim}

\begin{figure}
    \centering
    \includegraphics{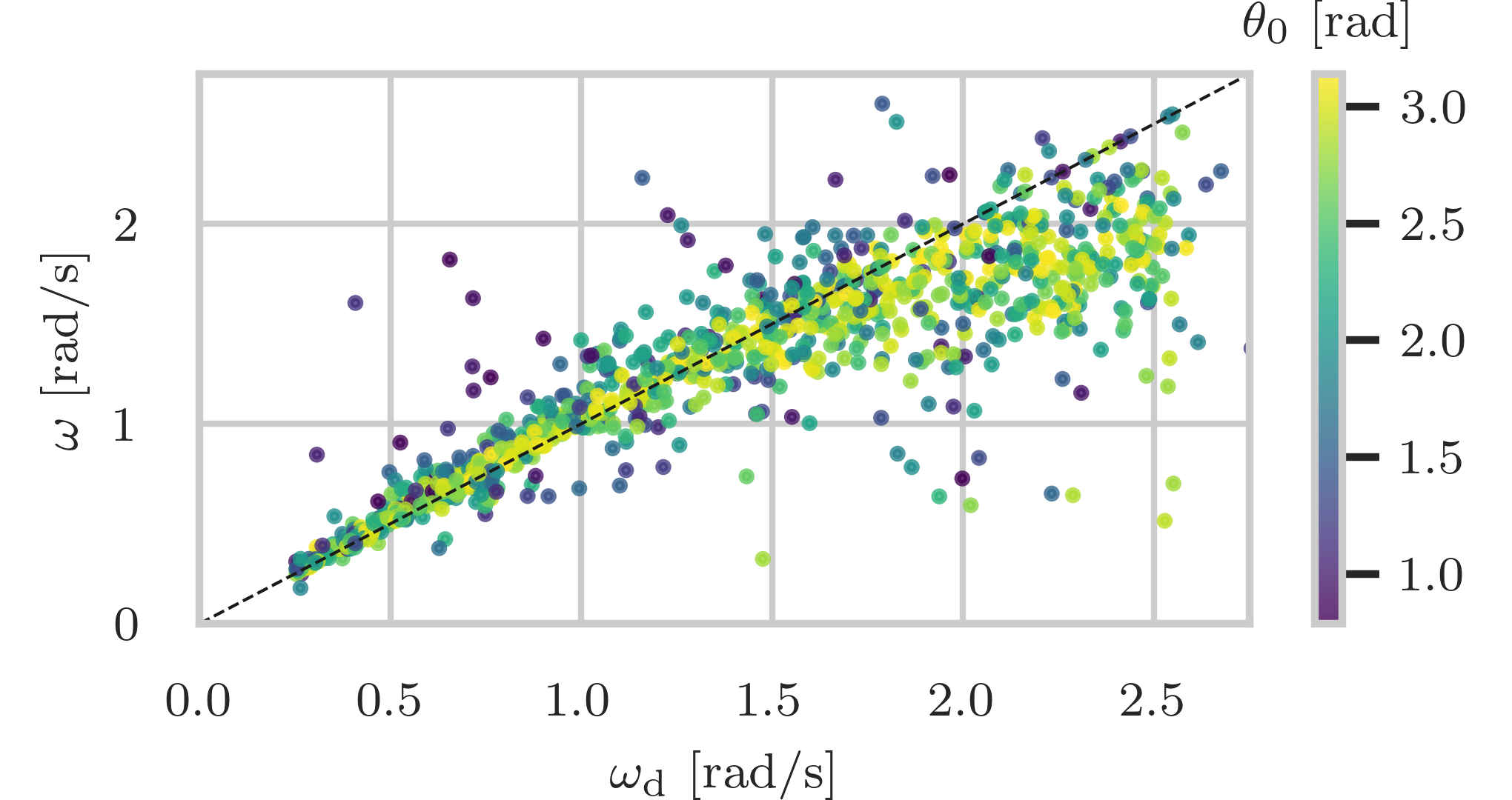}
    \caption{
        Scatter plot of the effective speed $\omega = \theta_0 / T$ against the target speed $\omega_\text{d}$ of individual evaluation episodes.
        The target speeds are sampled uniformly between 0.25 and $\SI{2.5}{rad/s}$, the same as during the training.
        We only plot successful trials (success rate $b =$ 93.5\%).
        The color indicates the initial angle $\theta_0$, showing no clear correlation between the initial angle $\theta_0$ and the difficulty for the policy to match the target speed.
        However, the variance increases significantly for small initial angles due to reorientations without regrasping.
    }
    \label{fig:scatter_speed}    
\end{figure}

In \cref{fig:scatter_speed}, we show how well the agent matches the target speed.
The correlation is excellent up to the speed of $\omega_\text{d}=\SI{1.5}{rad/s}$ and the curve saturates at around $\omega=\SI{2.0}{rad/s}$. 
This means, that the policy comes close to the ``time-optimal" maximum speed $\omega=\SI{2.5}{rad/s}$ from \cref{sec:time_optimal}.
However, it does not fully match it, indicating that the coupled training (EcRL) forces the policy to slow down.
This is a significant result since, besides the speed-adjustable reward, having the estimator in the loop poses a fundamentally more challenging learning problem.

Thereby, we were able to train an RL agent that can reorient cuboids with aspect ratios up to two from a known initial orientation to an arbitrary desired orientation with a configurable speed in the range from $0.25$ to about $\SI{1.5}{rad/s}$ with a high success rate in under 24 hours on a single Nvidia~T4 GPU.

\subsection{Evaluation on the Real System}

With experiments on the real DLR-Hand II, we validate that the results of the previous sections are attained in a realistic simulation.
We only run trials with a cube and use $\pi / 4$-discretized goals (cf. \cite{Pitz2023-ra}) to allow a human operator to determine success without requiring an object tracking system.
Episodes are stopped automatically when the estimated angle to the goal $\hat\theta_t$ is small enough for a few policy iterations, but the success of the episode and the time required to reach the goal are determined by examining the video footage, which is presented in the accompanying video.
Due to stress on the hardware caused by fast finger movements, we could not run enough trials to compare quantitatively with the simulation results.
However, we did validate all three full-$\pi$ rotations and two complex rotations (goal number 19 and 22 in the nomenclature of \citet{Pitz2023-ra}) at different target speeds $\omega_\text{d} = 0.5,\, 1.0 \text{ and } 2.0\,\text{rad/s}$.
Of those 30 trials, only one failed.
\cref{fig:title_figure} shows the successful reorientation for goal number 22 for the three different target speeds.

Despite the factor of four between the slowest and the fastest desired speed, the reorientation works robustly on the real hand. 
The execution times are difficult to read off exactly.
For a fair comparison with the simulation results, we need to judge from the video when the cube is within the goal threshold $\Delta_g$ and cannot rely on the manipulation to be stopped at that time.
For the few trials we checked carefully, we found that the required time closely matched the target speed ($<$ 10\% off).
We encourage the reader to inspect \cref{fig:title_figure} and, in particular, the accompanying video.
Especially for $\omega_\text{d} = \SI{2.0}{rad/s}$, this is a noteworthy result as it demonstrates the fastest complex in-hand manipulation task that was ever shown on a real robot without vision.

\section{Conclusion}

\label{sec:conclusion}

This paper presented methods for learning time-optimal and speed-adjustable in-hand reorientation policies.
Through the combination of a basic reward with a simple time-optimal objective, we were able to obtain policies that perform in-hand reorientations significantly faster than previous approaches while maintaining high success rates.
Moreover, we have shown that the policy can be trained to perform in-hand reorientations at variable speeds by appropriately conditioning the control policy on a target speed signal and optimizing a speed-parametrized objective.
We have demonstrated the effectiveness of the proposed methods in simulation and real-world experiments with the DLR-Hand~II, where we achieved robust, fast in-hand manipulation while using only tactile (via torque sensors) feedback.

\footnotesize
\bibliographystyle{IEEEtranN-modified}
\bibliography{IEEEabrv, bibliography}

\end{document}